\pdfoutput=1
\documentclass[11pt]{article}
\usepackage{subcaption}
\usepackage{wrapfig}
\usepackage{graphicx}
\usepackage{float}
\usepackage{microtype}
\usepackage{stfloats}
\usepackage[dvipsnames]{xcolor}
\usepackage{colortbl}
\definecolor{mybr}{RGB}{212,154,13}
\definecolor{myblue}{RGB}{46,108,164}
\usepackage[final]{acl}
\usepackage{amsmath}
\usepackage{listings}
\DeclareMathOperator*{\argmax}{arg\,max}
\usepackage{times}
\usepackage{amsfonts}
\usepackage{latexsym}
\usepackage{booktabs}
\usepackage{multirow}
\usepackage[T1]{fontenc}
\usepackage[utf8]{inputenc}

\usepackage{microtype}

\usepackage{inconsolata}

\title{EchoSight: Advancing Visual-Language Models with Wiki Knowledge}

\author{Yibin Yan \hspace{30pt} Weidi Xie\\[10pt]
School of Artificial Intelligence, Shanghai Jiao Tong University \\[3pt]
{\url{https://go2heart.github.io/echosight}}\vspace{-5mm}
}

\begin{document}
\maketitle
    \begin{figure*}[ht]
    \centering
    \includegraphics[width=1\textwidth]{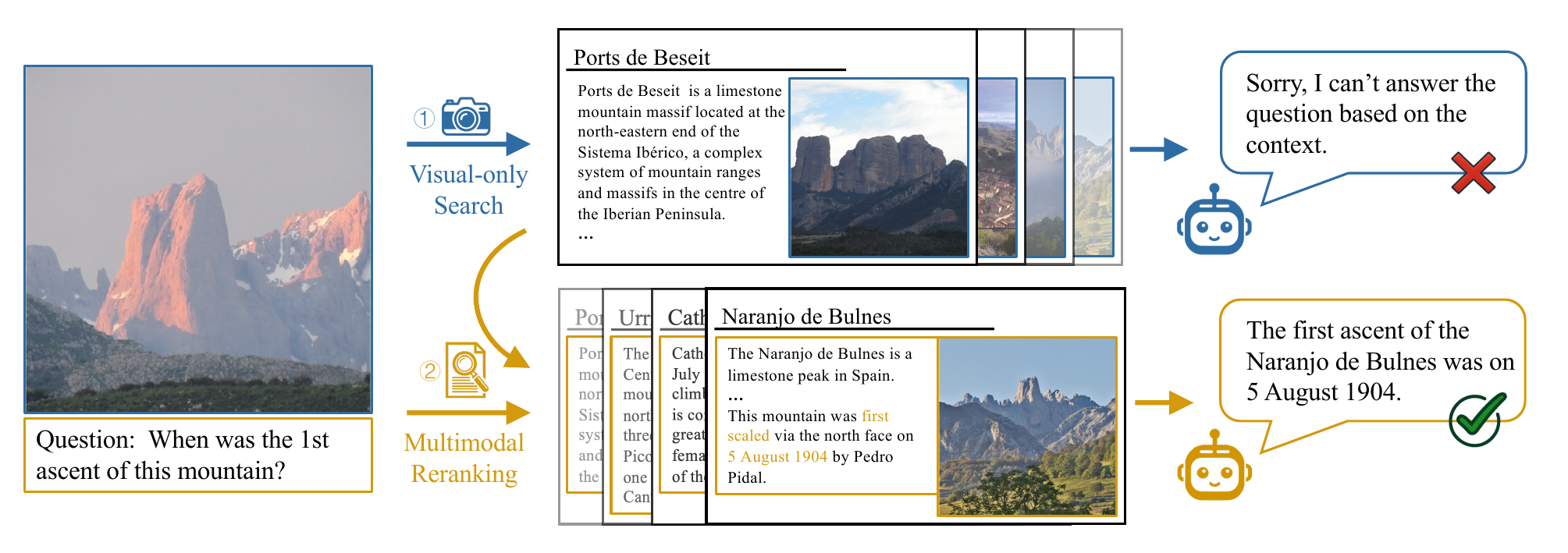}%\linewidth]{images/teaser3.pdf}
    \caption{For visual questions such as \textit{“When was the 1st ascent of this mountain?”}, {\color{myblue}visual-only search} methods consider image similarity only, ignoring the textual details of the accompanying article. By incorporating {\color{mybr}multimodal reranking}, the correct entry, accounting for both visual and textual information, can be accurately identified.}
    \label{fig:teaser}
\end{figure*}
\begin{abstract}

Knowledge-based Visual Question Answering (KVQA) tasks require answering questions about images using extensive background knowledge. Despite significant advancements, the large generative visual-language models often struggle with these tasks due to the limited integration of external knowledge. In this paper, we introduce \textbf{EchoSight}, a novel multimodal Retrieval-Augmented Generation (RAG) framework that enables to answer visual questions requiring fine-grained encyclopedic
knowledge. To strive for high-performing retrieval, EchoSight first searches wiki articles by using visual-only information, subsequently, these candidate articles are further reranked according to their relevance to the combined text-image query. This approach significantly improves the integration of multimodal knowledge, leading to enhanced retrieval outcomes and more accurate VQA responses. 
Our experimental results on the Encyclopedic VQA and InfoSeek datasets demonstrate that EchoSight establishes new state-of-the-art results in knowledge-based VQA, achieving an accuracy of 41.8\% on Encyclopedic VQA and 31.3\% on InfoSeek.
\end{abstract}

\section{Introduction}
% Imagine yourself standing before an encyclopedia. You are tasked with answering questions regarding its detailed content, including associated images. How would you proceed? The most straightforward method would be to quickly skim through the encyclopedia to identify sections where the images closely resemble those provided. Subsequently, closely examine the question and images to confirm if they are related to any specific article. Finally, organize the answer based on the article that contains the answer to complete the task. 
%\yibin{Changed for naming the Knowledge-based VQA at first}

Visual Question Answering (VQA) addresses the challenge of enabling machines to understand and respond to questions about visual content, typically images or videos. Broadly, this task can be divided into two categories: standard VQA~\cite{VQA,balanced_vqa_v2} with questions that can be answered directly from the visual content, for example, counting objects, identifying colors, or recognizing simple actions, which rely solely on commonsense and information present in the image; 
and knowledge-based VQA~\cite{marino2019ok,schwenk2022okvqa,mensink2023encyclopedic,chen2023can} requiring additional context or external knowledge, such as historical facts, detailed object properties, or specific situational contexts not evident in the visual content.
%Broadly, these questions may fall into two categories: those that can be answered directly from the visual content, for example, counting objects, identifying colors, or recognizing simple actions, which rely solely on commonsense and information present in the image; 
%and those that require additional context or external knowledge, such as historical facts, detailed object descriptions, or specific situational contexts not evident in the visual content.

Addressing these two types of questions presents different challenges for VQA systems. Questions that draw answers directly from visual content demand robust image understanding capabilities, encompassing tasks such as object detection, scene recognition, and spatial reasoning. Conversely, questions requiring external knowledge call for additional mechanisms to access and integrate information from external sources. In this paper, we focus on the latter type of visual question answering, by building a retrieval-augmented multimodal system, that enables searching an external knowledge base for more nuanced understanding and accurate responses.

Despite the recent accomplishments in developing Visual-language Models~(VLMs)~\cite{achiam2023gpt,team2023gemini,abdin2024phi,liu2024visual}, knowledge-based VQA remains challenging. 
This complexity primarily stems from two aspects.
(i) Existing VLMs struggle to adequately encode all essential knowledge, 
due to its limited model capacity, and infrequent inclusion of encyclopedic, long-tail information in their training data~\cite{mensink2023encyclopedic}.
(ii) The visual component of the questions often provides limited help in addressing the queries, as establishing a meaningful connection between entity knowledge and visual attributes can be difficult. 
For example, an image of a church alone barely reveal information about its construction date. 

%In this paper, we introduce EchoSight, a novel retrieval-augmented vision-language system for knowledge-based question answering. EchoSight features an integration of a retrieval-and-reranking search mechanism with the Retrieval Augmented Generation (RAG) paradigm. Specifically, EchoSight's search mechanism is a two-stage process, which comes with a visual-only retrieval, proceeded with a multimodal reranking. Initially, the visual-only retrieval from an external knowledge base ensures that the initial set of candidates closely matches to the visual context provided by the reference image. This stage narrows down the search space, focusing on visually similar images that are likely to be relevant. Subsequently, the reranking stage then attends these candidates by incorporating the reference image and textual question. This multimodal consideration ensures that the final results are not only visually relevant but also contextually relevant to the multimodal query. Once the coarse-to-fine grained search returns passages containing evident information to answer the question, our system then generates the desired answer.

In this paper, we introduce \textbf{EchoSight}, a novel retrieval-augmented vision-language system designed for knowledge-based visual question answering. 
EchoSight employs a dual-stage search mechanism that integrates a retrieval-and-reranking process with the Retrieval Augmented Generation (RAG) paradigm. Initially, the system performs a visual-only retrieval from an external knowledge base, to effectively narrow the knowledge search space,
only focusing on candidates that are closely align with the visual context of the reference image. In the subsequent multimodal reranking stage, the system refines the candidates ranking by incorporating both the reference image and the textual query. This approach guarantees that the selected results are pertinent not only visually, but also contextually to the multimodal query. After acquiring the most relevant information through this coarse-to-fine grained search, our model generates the precise answer to the posed question.

Overall, we present three contributions:
{\em First}, we propose a multimodal retrieval-augmented generation framework, termed as \textbf{EchoSight}, 
that enables to answer visual questions that require fine-grained encyclopedic knowledge;
{\em Second}, we adopt a retrieval-and-reranking scheme to improve retrieval performance, specifically, it first searches images with visual-only information, and then conduct a fine-grained multimodal reranking on the candidates;
{\em Third}, we conduct thorough experiments on both Encyclopedic VQA~\cite{mensink2023encyclopedic} and InfoSeek~\cite{chen2023can} benchmarks, \textbf{EchoSight} demonstrates state-of-the-art performance on both benchmarks, significantly outperforming existing VLMs or other retrieval-augmented architectures.

    \begin{figure*}[ht]
    \centering
    \includegraphics[width=1.0\textwidth]{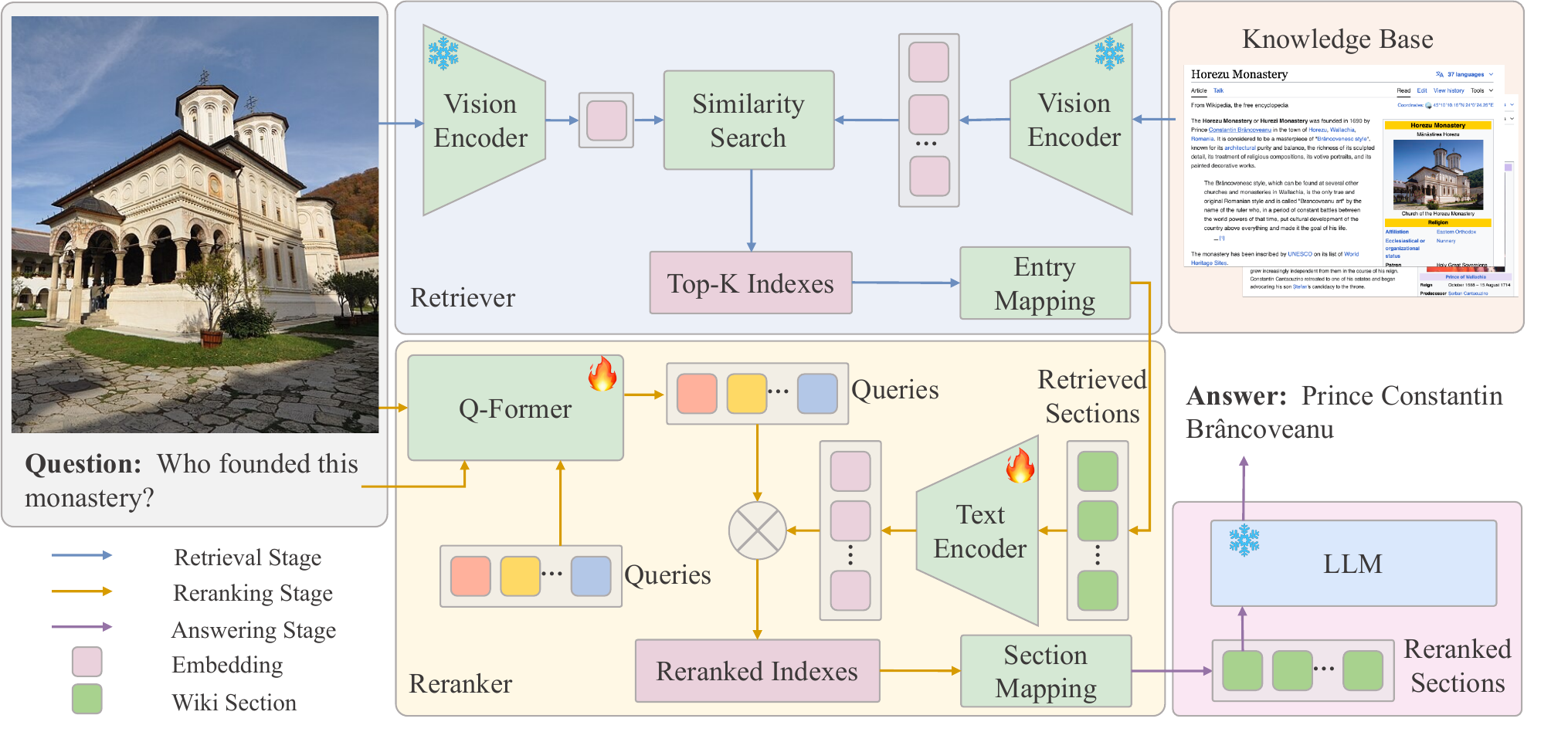}
    \caption{\textbf{The overall view of our proposed EchoSight}. \textbf{(i)} Given a visual question with an image, the retriever searches the reference image in the knowledge base for top $k$ similar images to get their corresponding Wikipedia Entries. \textbf{(ii)} After changing the granularity to sections, all the sections of retrieved entries are then reranked with the maximum pairwise similarity of their textual embeddings and the reference image+question's  Q-Former query tokens. \textbf{(iii)} The top reranked section will be utilized as RAG prompt for the LLM to generate the ultimate answer. }
    \label{fig:arch}
\end{figure*}

\section{Method}
This section starts with the problem formulation of retrieval-augmented VQA~(Sec.~\ref{sec:3.1}), followed by detailing the retrieval-and-reranking module in EchoSight~(Sec.~\ref{sec:3.2}), and finally the answer generation module~(Sec.~\ref{sec:3.3}).
% \weidi{Give 1-2 sentences on the high-level description of this section. `In this section, we start by giving the problem formulation of retrieval-augmented visual question answering~(vqa),
% then.... lastly, ...'}

\subsection{Problem Formulation}
\label{sec:3.1}
Given a reference image, and question of free-form texts, our goal is to construct a visual question answering system, that can benefit from the access of an external knowledge base. In our case, this is a million-scale dataset of entity articles and their corresponding images from Wikipedia webpage, 
{\em i.e.}, $\mathcal{B} = \{(a_1, I_1), \dots, (a_n, I_n)\}$. 
%\yibin{explain I \& a}
%\begin{align*}
%y = \Phi_{\text{rag-vqa}}(I_\text{ref}, q; \mathcal{B})
% \end{align*}
%where $y$ denotes the generated answers with free-form texts.

%Note that all entities presented in the multimodal queries will be included as article entries in the knowledge base $B$. The knowledge base is a million-scale dataset containing Wikipedia entity articles and their corresponding images.

The overall architecture of our proposed method, EchoSight, is illustrated in Figure \ref{fig:arch}. It consists of four main components: an external knowledge base (KB), a retriever, a reranker, and an answer generator. 
(i) The process begins with the retriever, which utilizes the reference image to filter and extract relevant KB entries with similar images;
(ii) Next, the reranker takes these candidate entries and employs their textual contents to have them reranked, based on their relevance to both the reference image and the textual question;
(iii) Finally, the reranked KB entries are fed into the answer generator to produce the final answer.

% In the following sections, we demonstrate our two-staged retrieve-and-rerank search pipeline. Finally, we describe the answer generation mechanism employed by our system.

%%% temp figure, to be changed later

\subsection{Retrieval and Reranking}\label{sec:3.2}
%The objective of this stage is to search the relevant entries from a large-scale external knowledge base, based on the given reference image and question. Specifically, we consider a two-stage procedure: (i) visual-only search, that discover candidates similar to the query image, solely relies on visual information; (ii) multimodal reranking, that evaluates both visual and textual information to rerank the retrieved entries, such that the desired article entry can be ranked in the first position to facilitate subsequent answer generation. 

The goal of this stage is to identify relevant entries from a large-scale external knowledge base using the given reference image and question. We employ a two-stage procedure: 
first, a visual-only search identifies candidates that are visually similar to the query image. 
Subsequently, a multimodal reranking process evaluates both visual and textual information to reorder the retrieved entries. This ensures that the most pertinent article entry can be ranked at the top, facilitating efficient and accurate answer generation.

\vspace{3pt}
\noindent \textbf{Visual-only Search.} 
Given the extensive size of the knowledge base, 
potentially encompassing millions of image-article pairs, 
optimizing the efficiency of the image search process is critical. 
To achieve this, we transform all images into vectors and utilize the cosine similarity metric to assess their proximity to a reference image. 
\begin{align*}
S_\Omega= \Biggl\{ s_i = \biggl \langle \frac{v_r}{||v_r||} \cdot \frac{v_i}{||v_i||} \biggr \rangle, i= 1, \dots, n \Biggl\},
\end{align*}
where $v_r= \Phi_\text{vis}(I_\text{ref})$ and $v_i = \Phi_\text{vis}(I_i)$ denote the visual embedding for reference image and database image, respectively, computed by a pre-trained visual encoder. We employ the FAISS library\cite{douze2024faiss} for vector search, and keep the top $k$ best-matched images and their corresponding wiki article entries from the knowledge base, {\em i.e.}, $\mathcal{E}_{v} = \{(a_1, I_1), \dots, (a_k, I_k)\}, k \ll n$.

\vspace{4pt} 
\noindent \textbf{Multimodal Reranking.}
After initially filtering the candidates based on visual similarities, 
the reranker module integrates both textual and visual inputs from the multimodal query and the top \( k \) retrieved Wikipedia article entries.
This stage aims to prioritize entries that are most pertinent to the question, ensuring the articles with highest relevancy are ranked at the top.

Specifically, we employ the Q-Former~\cite{li2023blip} architecture to extract multimodal information from the reference image and textual question, resulting $32$ query tokens. 
\begin{align*}
    z_m^i = \text{Q-Former}\left( I_{\text{ref}}, Q \right)^i,
\end{align*}
where \(z_m^i\) denotes the $i$th query token embedding of the reference image \(I_{ref}\) and textual question \(Q\).

On the candidates side, we break each of the wiki articles into sections, with each section prefixed by the article’s title, for example, $a_i = \{ \text{sec}_1^i, \text{sec}_2^i, \dots, \text{sec}_p^i \}$, and further encode them with Q-Former’s text encoder. 
We initialize the Q-Former with BLIP-2's weights and fine-tune all parameters except the visual encoder.

The reranking score for each section is calculated as follows:
\[
S_r^{\text{sec}} = \max_{1 \leq i \leq N_q} \left( \text{sim}(z_m^i, z_s^{\text{sec}}) \right),
\]
where \( S_r^{\text{sec}} \) is the reranking score for section “sec”, determined using the Q-Former’s Image-to-Text Correspondence (ITC) method. 
This method computes the highest pairwise similarity between each multimodal query token embedding \( z_m^i \) from the reference image and question pair, and the \texttt{[CLS]} token embedding of a Wikipedia article section \( z_s^{\text{sec}} \). \( N_q \) denotes the number of query tokens.

In the final step of multimodal reranking, 
the reranker combines the visual similarity score from the previous stage and the reranking score into a weighted sum:
\[
\text{sec}_{vl} = \argmax_{\text{sec} \in a} \left( \alpha \cdot S_v^{\text{sec}} + (1 - \alpha) \cdot S_r^{\text{sec}} \right),
\]
where \( \text{sec}_{vl} \) refers to the highest-ranked entry section produced by the reranker, \( \alpha \) is a weight parameter that balances the visual similarity score \( S_v^{\text{sec}} \) and the reranking score \( S_r^{\text{sec}} \). Note that, $S_{v}^{sec}$ is calculated in the visual-only search stage using the best-matched image from the wiki entry to which $sec$ belongs.

\vspace{3pt} \noindent \textbf{Reranker Training.}
Here, we implement hard negative sampling within a contrastive learning framework. Specifically, the negative samples are specifically selected from examples that are visually similar yet contextually distinct, 
{\em i.e.}, the initial visual-only retrieval efforts were unsuccessful.
With such training, the reranker is thus forced to select the most relevant articles for the multimodal queries, enhancing the overall accuracy and effectiveness of the system~\cite{robinson2021contrastive}.

The training objective of the reranker is given as follows:
\begin{align*}
\mathcal{L} = - \log \frac{\exp(\max_{1 \leq i \leq N_q} \text{sim}(z_{m}^i, z_{s}) / \mathcal{T})}{\sum_{j=1}^{N} \exp(\max_{1 \leq i \leq N_q} \text{sim}(z_{m}^i, z_{s}^j)/ \mathcal{T})},
% \mathcal{L} = - \log \left[
% \frac{\exp(\text{sim}(\mathbf{z}, \mathbf{z}_P))}
% {\exp(\text{sim}(\mathbf{z}, \mathbf{z}_P)) + \sum_{i=1}^{k} \exp(\text{sim}(\mathbf{z}, \mathbf{z}_{N_i}))}
% \right]
\end{align*}
where $z_s$ denotes the positive section embedding, $N$ is the total number of samples including both positive and negatives and $\mathcal{T}$ is the temperature parameter that controls the smoothness of the softmax distribution.

% \end{small}
%where the multimodal query token output $z_{m}$ is embedded by the Q-Former taking the reference image $I_{ref}$ and contextual question $q$ at the same time, and the Wikipedia section embedding is selected as the [CLS] token generated by feeding the Q-Former the section of Wikipedia atticle. Additionally, the embedding $z_{w}^j$ corresponds to the embedding of the $j$-th wiki entry section, which includes both positive and negative samples. 

%%% temp figure, to be changed later
% \begin{figure*}[h]
%     \centering
%     \includegraphics[width=0.8\textwidth]{images/retriver_arch.png}
%     \caption{The overall architecture of proposed method}
%     \label{fig:search_arch}
% \end{figure*}

\subsection{Answer Generation with LLMs}\label{sec:3.3}
Once the relevant entries are identified from the knowledge base,
large language models (LLMs) will integrate such information to answer the questions, {\em i.e.}, $A = \text{LLM}(\text{sec}_{vl}, Q)$, 
where the off-the-shelf LLM acts as an answer generator, 
$\text{sec}_{vl}$ denotes the retrieved wiki article section, 
and $Q$ refers to the target question. Comparing to existing generative VLMs, such retrieval-augmented generation~(RAG)~\cite{lewis2020retrieval}, enables the model with the essential contextual knowledge, improving the system's ability to handle complex questions that demand precise and detailed knowledge.

%EchoSight exploits this capability to enhance the accuracy and depth of answers, ensuring the system effectively handles complex queries that demand precise and detailed knowledge.

%Once the relevant knowledge base entries are searched out,  LLMs will integrate this information to generate answers. The generation of the final answer is achieved via Retrieval-Augmented Generation (RAG). By providing the LLM with crucial context containing knowledge of the multimodal query, the answers responded are significantly improved compared to vanilla generation. Modern LLMs are now capable of processing extremely long token sequences, allowing them to encapsulate more extensive information and intricate details from lengthy articles. This capability is leveraged by EchoSight to enhance the quality and accuracy of the answers, ensuring that the system can handle complex queries requiring detailed and specific knowledge.

% The generation of the final answer is achieved via Retrieval-Augmented Generation (RAG). By providing the LLM crucial context containing knowledge of the multimodal query, the answers generated will be greatly improved compared to vanilla generation. Modern LLMs are now capable of processing extremely long token sequences, thus allowing them to encapsulate more extensive information and intricate details from lengthy articles.

\section{Experiments}
\subsection{Datasets}
\noindent \textbf{Encyclopedic VQA~\cite{mensink2023encyclopedic}} contains 221k unique question and answer pairs each matched with (up to) 5 images, resulting in a total of 1M VQA samples.  These images are derived from iNaturalist 2021 (iNat21)~\cite{van2021benchmarking} and  Google Landmarks Dataset V2 (GLDv2)~\cite{weyand2020google}. The visual questions focus on the fine-grained categories and instances. There are single-hop and two-hop questions that require  different reasoning steps in the dataset. Notably, the dataset provides a controlled knowledge base with 2M Wikipedia articles with images, ensuring all the questions can be answered if correct Wikipedia article is given. 
For our experiments on E-VQA, we consider the single-hop questions using the provided 2M knowledge base. 

\vspace{3pt} 
\noindent \textbf{InfoSeek~\cite{chen2023can}}
comprises 1.3M visual information-seeking questions, 
covering more than 11K visual entities from OVEN~\cite{hu2023open}. 
InfoSeek provides a knowledge base with 100K Wikipedia articles with images. 
The questions of the dataset are diverse and the answers can be referenced from Wikipedia. There are a human-labeled 8.9K collection and an automated generated 1.3M collection in InfoSeek. Due to the unavailability of groundtruth for test split, we report evaluation results on the validation split. We note that, the original authors did not publicly release their knowledge base, 
we therefore filter a 100K knowledge base from E-VQA instead. 
We will release ours to the community for reproduction and future comparison.
%In the absence of the 100K knowledge base, we filter a 100K knowledge base from E-VQA. %After excluding the data whose Wikipedia article is absent in our 100K knowledge base, approximately 97\% (both training split and validation split) of the examples remained intact.

\subsection{Metrics}
To evaluate the performance of our proposed retrieval-augmented QA model, 
we focus on two aspects, namely, retrieval and question answering. 
The retrieval results gauge the system's capability to accurately retrieve relevant articles from a large-scale multimodal knowledge base, while the question answering results assess its holistic effectiveness in providing precise and correct answers to visual questions

\vspace{3pt}\noindent \textbf{Metrics for Retrieval.}
We utilize the standard metric Recall@K.  
Recall@K assesses whether the correct article entries appear among the top $k$ retrieved results. An article is considered correct only if its URL exactly matches the target URL, making our retrieval evaluation more stringent and precise compared to methods that only match the content of answers to the retrieved articles.

\vspace{3pt}\noindent \textbf{Metrics for Question Answering.}
Here, we follow the conventional practise, use different metrics depending on the considered datasets. For E-VQA dataset~\cite{mensink2023encyclopedic}, we use the BEM score~\cite{zhang2019bertscore}, while for the InfoSeek dataset~\cite{chen2023can}, we employ the VQA accuracy~\cite{balanced_vqa_v2,marino2019ok} and \textit{Relaxed Accuracy}~\cite{methani2020plotqa, masry2022chartqa}. These metrics are chosen to align with the evaluation settings specific to each dataset.

\begin{table}[t]
\small
\setlength{\tabcolsep}{8pt}
\begin{tabular}{lcccc}
\toprule
\multirow{2}{*}{Method} & \multicolumn{4}{c}{Recall@K}  \\
 & K=1 & K=5 & K=10 & K=20 \\
\midrule
Google Lens & 47.4 & 62.5 & 64.7 & 65.2 \\
CLIP I-T & 3.3 & 7.7 & 12.1 & 16.5 \\ \midrule
% PreFLMR & 12.1 & 21.8 & 26.7 & 31.6 \\
% CLIP I-I & 10.1 & 19.1 & 24.7 & 31.4 \\
\textbf{EchoSight}  \\
\hspace{0.5em}w/o. Reranking& {13.3} & {31.3} & {41.0} & {48.8} \\
\hspace{0.5em}w. Reranking&{\textbf{36.5}} & {\textbf{47.9}} & {\textbf{48.8}} &{48.8} \\
% \rowcolor{red}
% \hspace{0.5em}w. DinoV2&{\textbf{40.8}} & {\textbf{50.7}} & {\textbf{51.4}} &{51.4} \\
% RAVLM \\ (Rerank-100) & 18.3 & 32.4 & 39.2 & 45.1 \\
% RAVLM \\ (Sec Rerank) & \textbf{22.1} & 36.4 & 43.0 & 47.3\\
% RAVLM \\ (Text Rerank)  & \textbf{25.0} & 32.4 & - & - \\
\bottomrule
\end{tabular}
\caption{\textbf{E-VQA retrieval experiments.} While Google Lens can be recognized as a \textit{upperbound} in E-VQA, CLIP I-T indicates the retrieval from the reference image to Wikipedia entry texts with CLIP~\cite{radford2021learning}.}

\label{tab:evqa_retrieval}
\end{table}

\subsection{Implementation Details}
\vspace{3pt} 
\noindent \textbf{The Retriever.}
We compute the visual embedding for the reference images and images from database with a frozen Eva-CLIP vision encoder (Eva-CLIP-8B)~\cite{sun2024eva}.
The pooled last-layer embedding are used as the features for computing cosine similarity between images, with FAISS library.

%that are queried in a FAISS vector database to identify the top $k$ entries with the highest cosine similarity. 

%The vision model is kept frozen across all experiments. 
%This model extracts image features from the pooled embeddings of the last layer. 
%Upon embedding a reference image, these features are queried in a FAISS vector database to identify the top $k$ entries with the highest cosine similarity. 
% \weidi{what tech has been used in FAISS, vector quantisation or fast search ?}

\vspace{3pt} 
\noindent \textbf{The Reranker.}
The reranking module is initialized with pre-trained BLIP-2~\cite{li2023blip} weights using the LAVIS Library~\cite{li-etal-2023-lavis}. The number of query tokens $N_q$ is 32 and weighting parameter $\alpha$ is $0.5$. Instead of using in-batch contrastive learning, we employ hard negative sampling, where each positive sample is paired with $N = 24$ negative samples. 

In practise, a positive sample is constructed using the evidence section text from the corresponding Wikipedia article. While for negative samples, we perform a visual-only search on the reference images. Knowledge base entries with images that fail to match the reference images ranked within the top $k$ are selected as negative samples. During training, we randomly sample sections from these negative entries as well as from the non-evidence sections of the positive entries.
Note that, as only E-VQA dataset provides labeled evidence sections for all its training data, we train the reranker on this dataset, and directly use it on InfoSeek in a zero-shot manner.

We adopt OneCycleLR~\cite{smith2019super} scheduler, with AdamW~\cite{loshchilov2018decoupled} optimizer. 
We use learning rate $10^{-4}$, batch size $6$, 
and the negative samples per example being $24$. 
For training the reranker module with 900K examples in Encyclopedic VQA, 
150K steps require 40 hours on 1 Nvidia A100 (80G).

%Therefore, the results we report are based on the same E-VQA-trained checkpoint, making our inference on InfoSeek zero-shot. Empirically, although the amount of training data for these benchmarks is approximately the same, E-VQA's data contains more question templates and knowledge base entries (2M vs. 100K), resulting in more robust training on E-VQA.

%the absence of such sections in the InfoSeek dataset is addressed using a textual reranker model bge-reranker-v2-m3~\cite{chen2024bge}. 

%It is noteworthy that we find the checkpoint trained on E-VQA training data outperforms the one trained on InfoSeek data in both benchmarks. 

% \weidi{reword the procedure on hard negative sampling, for example, `for all training images, we conduct the visual-only search, for those with targeted images ranked .... we do ...'} 

\vspace{3pt} 
\noindent \textbf{The Answer Generator.}
We use Mistral-7B-Instruct-v0.2~\cite{jiang2023mistral} as the question generator for E-VQA and LLaMA-8B-Instruct~\cite{llama3modelcard} for InfoSeek.
% Our Retrieval-Augmented Vision-Language System comprises an ensemble of dynamically switchable modules. We explore various module configurations to optimize system performance. 

% For the retrieval component, we utilize the CLIP Vision Transformer model (CLIP-ViT-Large-Patch14). This model extracts image features from the pooled embeddings of the final layer. Upon embedding a reference image, these features are queried in a FAISS-backed vector database to identify the top $k$ entries with the highest cosine similarity.

% The Q-Former component in the reranking module is initialized with pretrained BLIP-2~\cite{li2023blip} weights. To enhance its adaptability to the reranking task, we implement hard negative sampling. Specifically, for a given positive article entry correlated with the reference image and associated question, negative samples are extracted from the list generated by the retriever, excluding any positive instances. This methodology significantly enhances the reranker's performance by compelling it to discern the intricate relationships between multimodal queries and corresponding article entries. 

% We also experiment different LLMs to see how they make a difference in the final answer generation stage.

\begin{table}[t]
\small
\setlength{\tabcolsep}{8pt}
\begin{tabular}{lcccc}
\toprule
\multirow{2}{*}{Method} & \multicolumn{4}{c}{Recall@K}  \\
 & K=1 & K=5 & K=10 & K=20 \\
\midrule
% PreFLMR & 12.1 & 21.8 & 26.7 & 31.6 \\
DPR$_{V+T}^{*}$ & 29.6 & - & - & - \\
CLIP I-T &32.0 &54.0  &61.6  &68.2  \\
\midrule
% CLIP I-T &  &  &  &  \\ %TODO 
\textbf{EchoSight}\\
\hspace{0.5em}w/o. Reranking&{45.6} & {67.1} &{73.0} & {77.9} \\
\hspace{0.5em}w. Reranking&{\textbf{53.2}} &{\textbf{74.0}} & {\textbf{77.4}} & {77.9} \\
% \rowcolor{red}
% \hspace{0.5em}w. DinoV2&{\textbf{38.2}} & {\textbf{60.0}} & {\textbf{64.1}} &{64.5} \\
% CLIP I-I & 9.4 & 25.1 & 33.9 & 41.7 \\
% RAVLM \\ (Rerank-20) & 16.3 & 34.8 & 40.5 & 41.7 \\
% RAVLM \\ (Sec Rerank) &  &  &  & \\
% RAVLM \\ (Text Rerank)  & &  & - & - \\
% RAVLM \\ (Text Rerank)  & 25.0 & 32.4 & - & - \\
% RAVLM \\ (Sec Rerank) & \textbf{29.5} & 44.1 & 50.0 & 61.1\\
\bottomrule
\end{tabular}
\caption{\textbf{InfoSeek retrieval experiments.} Note that, DPR$_{V+T}^{*}$~\cite{lerner2024cross} actually used an in-house 1.5M knowledge base. Its recall is calculated by answer matching (if the answer appeared in the retrieved text) instead of the absolute article matching we used.}

\label{tab:infoseek_retrieval}
\end{table}

\begin{table*}[ht]
\centering
\small
\setlength{\tabcolsep}{6mm}
\begin{tabular}{lllcc}
\toprule
{Method} & {LLM} & {Retrieval} & {E-VQA} & {InfoSeek} \\
% \multirow{2}{*}{Method} & \multirow{2}{*}{LLM} & \multirow{2}{*}{Retrieval} & \multicolumn{1}{c}{E-VQA} & \multicolumn{3}{c}{InfoSeek} \\
 % & & & Single-Hop&Unseen-Q&Unseen-E &All \\
\midrule
% \multirow{6}{*}{Oracle Retrieval} & PaLM & Subject C & 31.0 \\
% & PaLM & KB Article & 78.4 \\
% & Mistral-7B& Subject C & 29.4 \\
% & Mistral-7B& KB Article & 84.8\\
% & LLaMA3-8B& Subject C &32.0  \\
% & LLaMA3-8B& KB Article &84.6  \\

\rowcolor{gray!10} 
{Google Lens} & PaLM & KB Article & 48.0 & - \\
\rowcolor{gray!10} {Google Lens} & PaLM & KB Section & 48.8 & - \\
\midrule

\multirow{3}{*}{Vanilla} & PaLM & - & 19.7& 1.0 \\
 & Mistral-7B & - & 21.0 & 0.4\\
 & LLaMA3-8B & - & 18.7 & 2.4\\ 
BLIP-2 & Flan-T5XXL & - & 12.6 & 12.5 \\
LLaVA-1.5 & Vicuna-7B & - & 16.3 &  9.5 \\
Wiki-LLaVA & Vicuna-7B & KB Section & 21.8 &  28.9 \\
DPR$_{V+T}^{*}$& Multi-passage BERT & KB Section & 29.1&  12.4 \\
\midrule
{\textbf{EchoSight}}\\
\hspace{0.5em}w/o. Reranking & Mistral-7B | LLaMA3-8B\footnotemark[1]  & KB Article & 19.4 &27.7\\
% \hspace{1em}w. Rerank & Mistral-7B & KB Section & \textbf{39.7} &29.32 &27.84 &28.56\\
\hspace{0.5em}w. Reranking & Mistral-7B | LLaMA3-8B\footnotemark[1] & KB Section & \textbf{41.8}  &\textbf{31.3}\\
% \rowcolor{red} 
% \hspace{0.5em}w. Reranking(DinoV2) & Mistral-7B | LLaMA3-8B\footnotemark[1] & KB Section & \textbf{43.9}  &\textbf{31.3}\\
% EchoSight (w/o. Rerank) & Mistral-7B & KB Article & 19.4 & 27.7\\
% EchoSight (w. Rerank) & Mistral-7B & KB Section & \textbf{39.7} \\
% EchoSight (w. Rerank) & Mistral-7B & KB Article & \textbf{41.0} & \textbf{31.3}\\
% RAVLM-Retrieve \& Rerank(top 100) & Mistral-7B & KB Article & \textbf{33.6} \\
% RAVLM-Retrieve \& Sec Rerank & Mistral-7B & KB Section & \textbf{37.3} \\
% RAVLM-Retrieve \& Text Rerank& Mistral-7B & KB Article & \textbf{39.6} \\
\bottomrule
\end{tabular}

\caption{VQA Accuracy comparison with the SOTA methods. \colorbox{gray!10}{Google Lens} method is the closed source top performer. Vanilla method indicates the LLM directly generate answers with textual questions only. BLIP-2~\cite{li2023blip} and LLaVA\cite{liu2024visual} are strong vision language models yet with no retrieval augmented. Wiki-LLaVA\cite{caffagni2024wiki} and DPR$_{V+T}^{*}$\cite{lerner2024cross} are recent works focusing on retrieval-augmented answer generation. Our proposed \textbf{EchoSight} is reported without and with multimodal reranking. } 
\label{tab:method_comparison}
\end{table*}

\subsection{Results}
In this section, we present experimental results on the E-VQA and InfoSeek benchmarks. 

% \weidi{the dataset description belongs to the Dataset section.}
% \weidi{give an high-level description on what will be shown in this results section, 
% for example, we start by evaluating the effectiveness of retrieval module.}

% \weidi{The results section needs to be re-organised, one should be on `Ablation study', the other should be `Comparison to SOTAs'.}

% \weidi{In ablation study, `On Retrieval', (i) visual encoder, (ii) re-ranking, (iii) hard negative sampling or not, `On question generation', effect of oracle article, 
% different LLMs}

% \weidi{In Comparison to SOTAs, reformulate Table 3, get rid of the oracle retrieval.}

\vspace{3pt} 
\noindent \textbf{On Retrieval.}
The experiment results for the retrieval tasks across different configurations are detailed in Table \ref{tab:evqa_retrieval} and Table \ref{tab:infoseek_retrieval}. 
The CLIP I-T setting involves using CLIP for cross-modal similarity search, 
from the reference image to the Wikipedia article. 
The articles are represented as CLIP embedding of their title and descriptions.
The `Google Lens' refers to the approach used in Encyclopedic VQA~\cite{mensink2023encyclopedic}, where Google Lens indexes billions of images from the Internet, not limited to Wikipedia, to find and return the most closely matching images along with an entity prediction. 
The best corresponding knowledge base entry identified by Google Lens is considered as its retrieval results. Given its vast image index, which potentially includes the image from the test set and capability to associate images with relevant entities, Google's retrieval can be viewed as a \textit{top performer} in E-VQA retrieval. 

From both tables, we can draw the observation that, our proposed reranking module has shown to significantly improve the retrieval performance, for example, it improves Recall@1 from 13.3\% to 36.5\% on E-VQA benchmark, 45.6\% to 53.2\% on InfoSeek, largely bridging the gap towards the `Google Lens' top performer.

%\weidi{the table reference is incorrect, check this paragraph.}
%\yibin{Different Recall metrics, ours need to be recalculated}

%In Table \ref{tab:evqa_retrieval}, the Google Lens method refers to the approach used in the Encyclopedic VQA~\cite{mensink2023encyclopedic}, where Google Lens is employed for retrieval purposes, and considered as an upperbound. Indexing billions of images on the Internet, Google Lens searches images beyond Wikipedia and returns the most similar indexed images along with an entity prediction. The best matching knowledge base entry for the entity is recognized as Google Lens retrieval result. By leveraging its extensive index of billions of images across the internet, Google Lens is capable of identifying highly similar images and associating them with relevant entities. Therefore, Google Lens retrieval can be regarded as \textit{soft-oracle} setting in E-VQA retrieval. 

%\weidi{unclear how Google Lens is applied, clarify it, for example, by tagging the image, and matching corresponding entities from the Wiki, blabla...}

\vspace{3pt} \noindent \textbf{VQA Results.} 
As shown in Table \ref{tab:method_comparison}, 
we present the comparison to state-of-the-art approaches on final VQA results.
For methods that do not utilize an external knowledge base or retrieval system, 
we present the results of large language models (LLMs), and multimodal large language models (MLLMs). The vanilla method refers to scenarios where only the textual question of the multimodal query is provided. The performance of multimodal-LLMs, including BLIP2~\cite{li2023blip} and LLaVA~\cite{liu2024visual}, are reported in Wiki-LLaVA~\cite{caffagni2024wiki}, where both the reference image and question are simultaneously processed. For methods with external knowledge bases, we compare with Wiki-LLaVA~\cite{caffagni2024wiki} and DPR$_{V+T}^{*}$~\cite{lerner2024cross}. 

It is clear that our proposed EchoSight~(w.~reranking) has outperform the prior works by a significant margin, even approaching the upperbound results reported by original E-VQA~\cite{mensink2023encyclopedic} benchmark, 
where two giant models are adopted, 
{\em i.e.}, `Google Lens' for knowledge retrieval, and PaLM for answer generation.

%from combination giant models, {\em i.e.}, `Google Lens + PaLM'.
%Specifically, the first two rows refer to the reported results from the original E-VQA~\cite{mensink2023encyclopedic} benchmark, where `Google Lens' is used for knowledge retrieval, and PaLM as answer generation.

%`Google Lens + PaLM' denotes the setting that uses Google Lens as the Retriever and PaLM for answer generation. The result is reported in E-VQA~\cite{mensink2023encyclopedic}.

% Oracle Retrieval indicates that the LLM always get the correct retrieval results, varying from title only to the whole article. Google Lens is the setting that uses Google Lens as the Retriever. Google Lens is an image retrieval system which indexes a huge amount of web images. Consequently, it can be considered a soft oracle for EVQA. 

\subsection{Ablation Study}
For all ablation studies, we use the E-VQA dataset. On the retrieval side, we conduct the following ablations: (i) to compare different vision backbones in retrieval,
(ii) to study the impact of reranking scope, and (iii) to investigate the importance of hard negative sampling. On final answer generation, we carry out ablation studies on: (i) the impact of various language models and (ii) to experiment with the answer generator under oracle settings.

\noindent \footnotetext[1]{The E-VQA accuracy is tested with Mistral-7B and InfoSeek accuracy is tested with LLaMA3-8B.}

% \end{table}
% }
% \subcaption{LLMs Accuracy}
% \label{ablation_1}
% \end{subtable}
% \hfill
% \begin{subtable}{0.52\linewidth}

\begin{table}[ht]
\small
\setlength{\tabcolsep}{3mm}
\begin{tabular}{lcccc}
\toprule
\multirow{2}{*}{Backbone} & \multicolumn{4}{c}{Recall@K}  \\
 & K=1 & K=5 & K=10 & K=20 \\
\midrule
\textbf{OpenAI-CLIP}  & & & &  \\
\hspace{0.5em}w/o. Reranking  &10.1 &19.5 &25.8 &32.2 \\
\hspace{0.5em}w. Reranking  &\textbf{23.8} & \textbf{31.4} &\textbf{32.1} &32.2 \\
\textbf{Eva-CLIP}  & & & &  \\
\hspace{0.5em}w/o. Reranking  &13.3 &31.3 &41.0 &48.8 \\
\hspace{0.5em}w. Reranking  &\textbf{36.5} &\textbf{47.9} &\textbf{48.8} &48.8 \\
\bottomrule
\end{tabular}
\caption{Retrieval performance analysis on different vision backbones.OpenAI-CLIP is CLIP-ViT-Large~\cite{radford2021learning} and Eva-CLIP is Eva-CLIP-8B~\cite{sun2024eva} from BAAI. We both take the visual encoder's last layer output as the image feature.}
\label{ablation_2}
\end{table}

% \resizebox{!}{1.3cm}{
% \begin{tabular}{ccc}
% \toprule
% Recall  & CLIP & EvaCLIP \\
% \midrule
% @1    & 10.1   & 13.3 \\
% @5    & 19.1   & 31.3 \\
% @10 & 24.7    & 41.0\\
% @20  & 31.4    & 48.8\\
% \bottomrule
% \end{tabular}
% }
% \subcaption{Recall Comparison}
% \label{ablation_2}
% % \end{subtable}
% \caption{(a) Accuracy of different LLMs in E-VQA, providing with EchoSight Retrieval-Reranking KB entries. The accuracy is calculated with BEM. (b) Recall comparison between CLIP and EvaCLIP at different recall levels. CLIP is OpenAI's clip-vit-large-patch14 and EvaCLIP is EvaCLIP-8B from BAAI. We both take the visual encoder's last layer output as the image feature.}
% \label{Ablation_combined}
% \end{table}

\vspace{3pt} 
\noindent \textbf{Impact of vision backbones.}
We assess the effect of different vision backbones on the retrieval stage, as detailed in Table \ref{ablation_2}. We compare the Vision Transformer (ViT) from EvaCLIP-8B~\cite{sun2024eva} with OpenAI's CLIP-ViT-Large~\cite{radford2021learning}. The EvaCLIP-8B's ViT achieves a recall@20 of 48.8\%, outperforming the CLIP-ViT-Large, which scored 32.2\%. 
This substantial improvement is likely due to EvaCLIP-8B's larger parameter size and more extensive training dataset, allowing it to develop more robust representations.

While the initial Recall@1 shows a modest difference between the two models (10\% for CLIP-ViT-Large and 13\% for EvaCLIP-8B), adopting our multimodal reranking significantly boosts performance, increasing Recall@1 to 23.8\% and 36.5\% for CLIP-ViT-Large and EvaCLIP-8B, respectively. This results in a marked 13\% difference, underscoring the effectiveness of our approach, especially when combined with a more capable backbone.

%Although the Recall@1  for the two backbone models initially shows a modest difference of 3\% (10\% vs. 13\%), the adoption of our multimodal reranking significantly enhances their performance. Specifically, our method boosts Recall@1 to 23.8\% and 36.5\% for the respective backbones. This results in an impressive 13\% difference, demonstrating the superior efficacy of our approach, particularly when integrated with a more robust vision backbone.

%These findings underscore the importance of model size and training data volume in enhancing retrieval performance, suggesting that future work should consider these factors when developing and selecting vision backbones.

\vspace{3pt} \noindent \textbf{Impact of reranking scope.}
%\yibin{TODO:To change according the new result}
The reranking scope refers to the number of candidates considered by the reranker module. Involving a higher reranking scope means calculating more embeddings during the reranking process. The reranking scope, which can be any number up to $k$, 
{\em i.e.}, the total number of candidates returned by the retriever. 
As shown in Table \ref{tab:rerank_scope}, our reranker can consistently improve the results with increasing scope from Top-5 to Top-500. As the thoughput experiment showed in Table \ref{tab:throughput}, considering the balance of efficiency and quality, the scope of 20 candidate entries is used when reporting our final VQA accuracy on E-VQA and InfoSeek.

\begin{table}[!htb]
\small
\setlength{\tabcolsep}{4mm}
\centering
\begin{tabular}{lcccc}
\toprule
\multirow{2}{*}{Scope} & \multicolumn{4}{c}{Recall@K}  \\
 & K=1 & K=5 & K=10 & K=20 \\
\midrule
Top 5 & 29.4& 32.2&-& -\\
Top 10 &34.3 &40.7 &40.9 &- \\ 
Top 20 & 36.5 & 47.9 & 48.8 & 48.8 \\
Top 50 & 38.3& 53.6& 56.9& 57.9\\
Top 100& 38.8 &55.9 &60.8 & 63.0\\ 
Top 500& 39.8 &58.5 &65.3 & 70.3\\
\bottomrule
\end{tabular}
\caption{The ablation study on impact of the reranking scope. 
Our reranker can consistently improve the results with increasing scope from Top-5 to Top-500.}
\label{tab:rerank_scope}
\end{table}
\begin{table}[!htb]
\small
\setlength{\tabcolsep}{4mm}
\centering
\begin{tabular}{lcccc}
\toprule
Scope & Total Retrieval Time & Throughput \\ \midrule
Top 10 & 0.602             & 1.66             \\
Top 20 & 1.171           & 0.85             \\
Top 50 & 2.720           & 0.37             \\
Top 100& 5.082           & 0.20             \\
Top 500& 21.591          & 0.05             \\
\bottomrule
\end{tabular}
\caption{Throughput Study. This study uses a NVIDIA A100 GPU with 80GB memory limiting the batch size to 1. For visual-only retrieval, we use the Faiss library with an exhaustive search. The throughput is calculated as the number of queries processed per second (QPS).}
\label{tab:throughput}
\end{table}

\vspace{3pt}

\noindent \textbf{Impact of hard negative sampling.}
The training strategy of the reranker module is critical for its performance. 
Rather than using randomly selected, irrelevant article entries, 
we employ a hard negative sampling during training, {\em i.e.}, top negative candidates returned by the retriever. This approach ensures the reranker to be trained on more demanding examples, thereby improving its performance and robustness. The effects of different training strategies on reranking performance are detailed in Table~\ref{tab:ablation_hard_neg}.

\begin{table}[!htb]
\small
\setlength{\tabcolsep}{9pt}
\begin{tabular}{lcccc}
\toprule
\multirow{2}{*}{Sampling} & \multicolumn{4}{c}{Recall@K}  \\
 & K=1 & K=5 & K=10 & K=20 \\
\midrule
\textbf{EchoSight}  & & & &  \\
\hspace{0.5em}w/o. Hard Neg  &31.4 &46.0 &48.5 &48.8 \\
\hspace{0.5em}w. Hard Neg  &\textbf{36.5} &\textbf{47.9} &\textbf{48.8} &48.8 \\
\bottomrule
\end{tabular}
\caption{The ablation study of how sampling methods affect the overall retrieval-and-reranking performance. }

\label{tab:ablation_hard_neg}
\end{table}

% \begin{wrapfigure}{l}{0.45\linewidth}

% \centering
% \begin{tabular}{cc}
% \toprule
% LLMs  & Accuracy \\

% \midrule
% GPT4    & 44.4    \\
% PaLM    & 39.0     \\
% Mistral & 42.6     \\
% LLaMA3  & 38.9   \\
% \bottomrule
% \end{tabular}
% \caption{Impact of language models.}
% \label{ablation_1}
% \end{wrapfigure}
\vspace{3pt} 
\noindent \textbf{Consistency of EchoSight across LLMs.}
The choice of LLMs influences the RAG paradigm greatly~\cite{shao2023prompting, hu2022promptcap}. We compare PaLM~\cite{chowdhery2023palm}, GPT-4~\cite{achiam2023gpt}, Mistral-7B-Instruct-v0.2~\cite{jiang2023mistral} and LLaMA3-8B-Instruct~\cite{llama3modelcard} as answer generators.
Specifically,
we provide them with same reranking results (KB entries). 
As shown in Table \ref{ablation_1}, the accuracy results are calculated with BEM~\cite{zhang2019bertscore} following~\cite{mensink2023encyclopedic}. 
The results indicate that though better language models yield better scores, the overall performance across all tested language models is quite stable. This validates our method adapts well across modern language models. 
\begin{table}[t]
    \centering
    \small
    \tabcolsep=0.19cm
    \begin{tabular}{ccccc}
    \toprule
    LLMs  & GPT-4 & PaLM & Mistral &LLaMA3\\
    \midrule
    Accuracy& 44.4  & 39.0 & 41.8 & 38.9   \\
    \bottomrule
    \end{tabular}
    \caption{The ablation study of impact of language models. The results are generated with the retrieval results of EchoSight with reranking scope 20.}
    \label{ablation_1}
\end{table}

\vspace{3pt} 
\noindent \textbf{Effect of oracle retrieval.}
Oracle retrieval indicates that the correct Wikipedia entry is always provided for generating the answer. As shown in Table~\ref{tab:ablation_orcale_vqa}, LLMs can \textit{almost}  answer the question if oracle retrieval is provided.

\begin{table}[ht]
\small
\centering
\setlength{\tabcolsep}{16pt}
\begin{tabular}{llc}
\toprule
LLM & Retrieval & Accuracy  \\
\midrule
PaLM & KB Title & 31.0 \\
Mistral-7B& KB Title & 29.4 \\
LLaMA3-8B& KB Title &32.0  \\ \midrule
PaLM & KB Article & 78.4 \\
Mistral-7B& KB Article & 84.8\\
LLaMA3-8B& KB Article &84.6  \\\midrule
PaLM & KB Section & 87.0 \\
Mistral-7B& KB Section & 89.9 \\
LLaMA3-8B& KB Section & 90.7  \\
\bottomrule
\end{tabular}
\caption{The ablation study with VQA results on the effect of oracle retrieval.}
\label{tab:ablation_orcale_vqa}
\end{table}

%performance with the top-100 retrieved candidates for the E-VQA dataset and its associated knowledge base. Adjusting the scope either below or above this range leads to a decline in performance. This underscores that the reranker functions most effectively within a particular candidate range, emphasizing the importance of setting an appropriate reranking scope to maximize efficacy.

\subsection{Qualitative Results}
\begin{figure*}[htbp]
    \centering
    \includegraphics[width=1.0\textwidth]{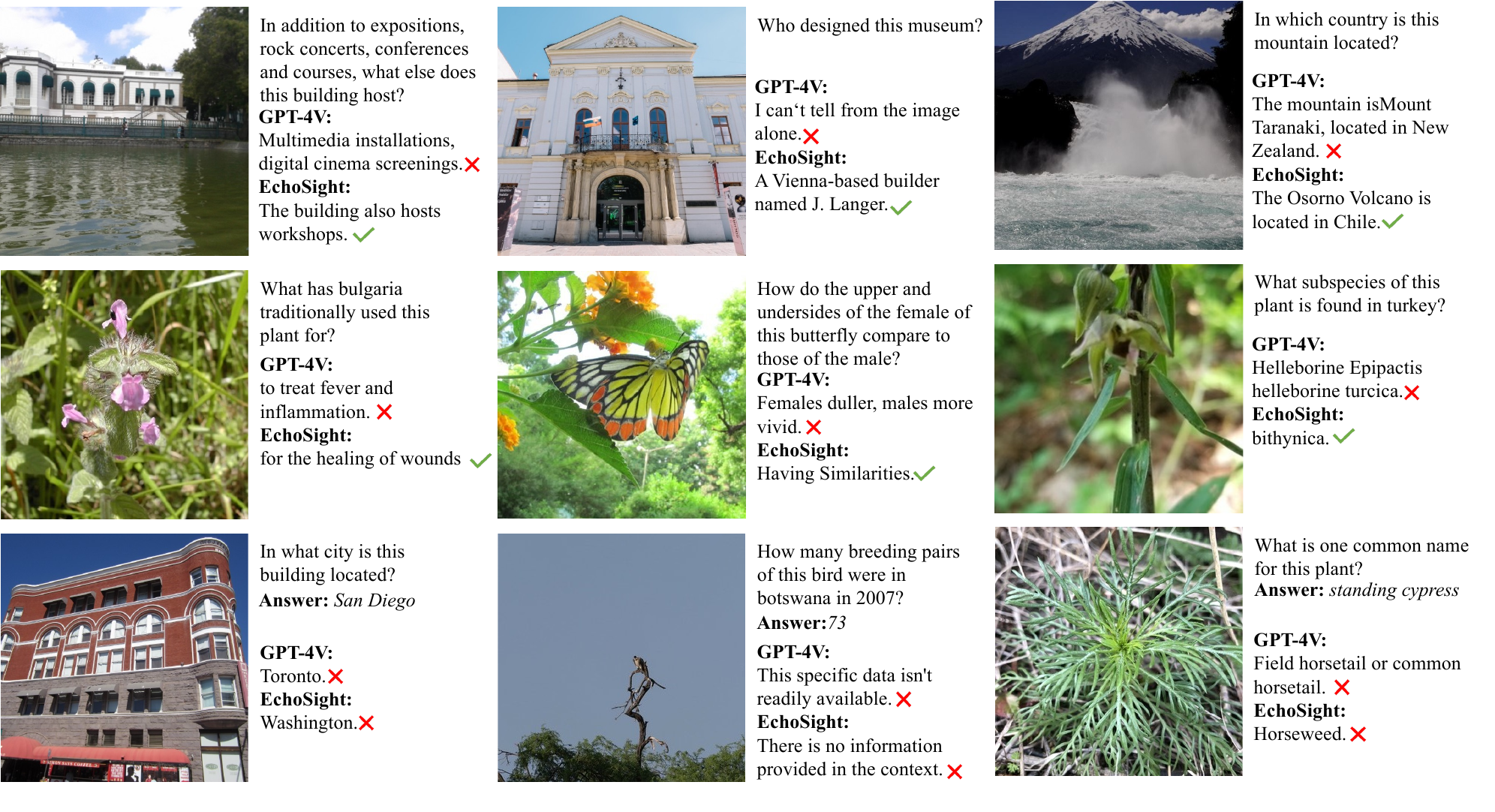}
    \caption{Qualitative VQA results from Encyclopedic VQA comparing to GPT-4V. The first row shows results in landmarks and the second row in natural species. Some failure cases are shown in the third row altogether with ground-truth.}
    \label{fig:qualitive_result_vqa}
\end{figure*}

As shown in Figure \ref{fig:qualitive_result_vqa}, our EchoSight demonstrates significant improvements in multimodal understanding and generation tasks compared to the state-of-the-art GPT-4V~\cite{achiam2023gpt}.

%The reranking scope refers to the number of candidates being considered by the reranker module. A higher reranking scope involves calculating more embeddings during the reranking process. The reranking scope can be any number less than $k$, the total number of candidates returned by the retriever. As shown in Table \ref{tab:rerank_scope}, for the E-VQA dataset and the corresponding knowledge base, our reranker performs optimally when considering the top-100 retrieved candidates. If the scope is reduced or expanded beyond this range, the reranker's performance deteriorates. This indicates that the reranker operates most effectively within a specific candidate range, highlighting the importance of selecting an appropriate reranking scope for optimal performance.
% \vspace{3pt} 
%\noindent \textbf{The Limit of reranker} TODO

    \section{Related Work}

% \weidi{Three subsections, one is to review the VQA, two types, standard VQA, knowledge-based VQA, etc.}
% \weidi{The other one is to review the VLM models, LLaVA, GPT-4V, Palm, etc, they are good at dealing with VQA, based on the information within visual signals, however, suffer from hallucination or other issues on knowledge-based VQA.
% Another line of research focuses on retrieval-augmented system, for example.....}
% \weidi{need a subsection on image retrieval.}

\subsection{Visual Question Answering}
Visual Question Answering (VQA) is the task of answering open-ended questions based on an image with natural language response. VQA tasks can be divided into two types: standard VQA and knowledge-based VQA.

\vspace{3pt}  
\noindent \textbf{Standard VQA.}
Datasets such as VQAv1~\cite{VQA}, VQAv2~\cite{balanced_vqa_v2}, and VizWiz~\cite{gurari2018vizwiz} focus on questions that can be answered by analyzing the image content alone, without external information. These datasets typically cover questions about objects in the image, their attributes and other perceptual details that can be inferred from the visual input.

\vspace{3pt}  
\noindent \textbf{Knowledge-based VQA.}
The task involves questions that require information not present in the image. Pioneering datasets like OK-VQA~\cite{marino2019ok} and A-OKVQA~\cite{schwenk2022okvqa}, which include questions needing knowledge beyond what is visually depicted, necessitate the integration of external world knowledge and commonsense reasoning. Uni-modal knowledge bases like GS112K~\cite{luo-etal-2021-weakly} and Wiki21M~\cite{karpukhin-etal-2020-dense} are adopted in prior works~\cite{lin2023fine, lin-byrne-2022-retrieval, gao2022transform, luo-etal-2021-weakly, luo2023end}. 
However, uni-modal knowledge bases are text-only, 
which limits their applicability in scenarios where visual context is paramount. To better utilize multimodal information, multiple previous attempts have been made~\cite{ding2022mukea, zhu2020mucko, wu2022multi, chen2022murag}.

% However, both datasets focus primarily on commonsense and general world knowledge, often neglecting more specialized or encyclopedic facts, and they do not provide external knowledge bases.

% % One of the pioneering datasets in this category is OK-VQA~\cite{marino2019okvqa}, which includes questions that necessitate knowledge beyond what is visually depicted, such as cultural, historical, or commonsense information. Similarly, A-OKVQA~\cite{schwenk2022aokvqa} extends this concept by introducing more complex and diverse questions, further emphasizing the need for robust external knowledge integration. However, both datasets primarily focus on commonsense and general world knowledge, often neglecting more specialized or encyclopedic facts, and they do not provide explicit external knowledge sources.

% Pioneering datasets like OK-VQA~\cite{marino2019okvqa} and A-OKVQA~\cite{schwenk2022aokvqa}, which include questions needing knowledge beyond what is visually depicted, necessitate the integration of external world knowledge and commonsense reasoning. However, both datasets focus primarily on commonsense and general world knowledge, often neglecting more specialized or encyclopedic facts, and they do not provide external knowledge bases.

Datasets such as Encyclopedic VQA (E-VQA)~\cite{mensink2023encyclopedic} and InfoSeek~\cite{chen2023can} have been developed with multimodal knowledge bases. These datasets utilize Wikipedia as a multimodal knowledge base to provide detailed and specific information on various topics. E-VQA covers a wide range of topics like animals, plants, and landmarks, while InfoSeek focuses on info-seeking questions about various visual entities. These datasets require models to recognize visual entities and accurately retrieve and use relevant information from external sources~\cite{lerner2024cross,caffagni2024wiki,lin2024preflmr}.

\subsection{Vison Language Models for VQA}

Advances in Vision Language Models (VLMs) such as GPT-4V~\cite{achiam2023gpt}, Gemini~\cite{team2023gemini}, LLaVA~\cite{liu2024visual}, and Phi-3-Vision~\cite{abdin2024phi} have demonstrated impressive capabilities in standard Visual Question Answering (VQA) tasks, exhibiting strong image analysis and accurate response generation~\cite{li2023comprehensive}. However, these models encounter difficulties with knowledge-based VQA due to issues such as hallucination, where responses are generated based on nonexistent content and internal biases~\cite{li2023evaluating}, and the lack of efficient knowledge retrieval mechanisms which hampers the integration of external knowledge bases for reasoning~\cite{caffagni2024wiki}.

Recently, research has shifted towards retrieval-augmented generative systems. While Retrieval-Augmented Generation (RAG) has been well-established in Large Language Models (LLMs), its application in VLMs remains underexplored. Systems like KAT~\cite{gui2021kat}, REVIVE~\cite{lin2022revive}, and REVEAL~\cite{hu2023reveal} show promise for questions involving commonsense reasoning, yet they struggle with complex, knowledge-intensive tasks like Encyclopedic VQA (E-VQA) and Infoseek. 
These limitations stem from their restricted ability to fetch and incorporate precise information from extensive encyclopedic knowledge bases~\cite{mensink2023encyclopedic}.

EchoSight addresses these issues through a novel two-stage process combining visual-only retrieval and multimodal reranking. This approach significantly enhances the alignment between retrieved textual knowledge and visual content, leading to improved performance on benchmarks such as Encyclopedic VQA and InfoSeek.

    \section{Conclusion}
In this paper, we introduced EchoSight, a novel retrieval-augmented vision language system designed to address the challenges of knowledge-based Visual Question Answering (VQA). Our approach enhances the retrieval capabilities of multimodal models through a two-stage process: initial visual-only retrieval followed by a multimodal reranking stage. This methodology significantly improves the alignment between visual and textual information, leading to more accurate and contextually relevant answers. Experimentally, we have conducted thorough ablation studies to demonstrate the effectiveness of our proposed components. 
While comparing to existing state-of-the-art approaches on the Encyclopedic VQA and InfoSeek datasets, EchoSight demonstrates significant performance improvement, with an accuracy of 41.8\% on E-VQA and 31.3\% on InfoSeek.
The success of EchoSight highlights the importance of efficient retrieval processes and the integration of multimodal information in enhancing the performance of large language models (LLMs) in knowledge-based VQA tasks.

    \section*{Limitations}
Although our proposed EchoSight demonstrates impressive performance on Knowledge-based VQA like Encyclopedic-VQA and InfoSeek, several limitations must be acknowledged. EchoSight's performance is heavily dependent on the quality and comprehensiveness of the underlying knowledge base used for retrieval. Domain-specific knowledge not covered in these databases may lead to sub-optimal performance in specialized queries. In addition, the retrieval process, especially when involving multimodal reranking of candidates, introduces significant computational overheads, making it less suitable for real-time applications. These overheads can impact the efficiency and response time of the system. Future work focusing on improving the quality of knowledge bases and mitigating computational overheads remains to be explored.

\section*{Acknowledgements}
This work is funded by National Key R\&D Program of China (No.2022ZD0161400).

\bibliography{main}

\clearpage
\appendix
\section{Dataset Details}

In this section, we provide more details of in the Dataset we used. We summarize the statistics of  in Table \ref{tab:evqa_stat}. 

\subsection{E-VQA}
We focus only on Single-hop questions of E-VQA~\cite{mensink2023encyclopedic}, namely Templated, Automatic, and Multi Answer questions in the table.

\begin{table*}[ht]
\centering
\setlength{\tabcolsep}{6mm}
\begin{tabular}{lcrrr}
\toprule
\multirow{2}{*}{Dataset} & \multirow{2}{*}{Question Type} & \multicolumn{3}{c}{Number of IQA pairs} \\ \cmidrule{3-5}

& & Train  & Val  & Test \\
\midrule
\multirow{4}{*}{E-VQA} 
& \multicolumn{1}{r}{Templated}  & 66,535        & 1,827        & 1,000       \\
& \multicolumn{1}{r}{Automatic}    & 737,114 & 8,025 & 2,750 \\
& \multicolumn{1}{r}{Multi Answer} & 112,736 & 1,844 & 1,000 \\ \cmidrule{2-5}
& \multicolumn{1}{l}{\textbf{Total}} &\textbf{916,385} & \textbf{11,696}& \textbf{4,750}\\
\midrule
InfoSeek & \multicolumn{1}{l}{\textbf{Total}}  &\textbf{902,509}& - & \textbf{71,335}    \\
\bottomrule
\end{tabular}
\caption{Dataset details used in our EchoSight's traning and testing.}
\label{tab:evqa_stat}
\end{table*}

\subsection{InfoSeek}
And for Infoseek~\cite{chen2023can}, due to the missing entities in the knowledge-base we use, we remove the examples in the dataset. Specifically, 916,385 examples in training split out of 934,048 are kept (98.1\%), and 71,335 examples of validation split out of 73,620 are kept (96.9\%). Therefore, the results we obtain with our knowledge base are consistent with the dataset's original setting while considering for the limitations of our knowledge base.

\section{Vision backbones}
In addition to CLIP, there are other robust vision backbones available, such as Dino~\cite{caron2021emerging,oquab2023dinov2}. Unlike CLIP, which employs a visual-language training method, Dino leverages a self-supervised, visual-focused training approach. To evaluate its performance, we benchmarked DinoV2 as our visual-only retriever, presenting the results in Tables \ref{tab:dino_evqa} and \ref{tab:dino_info}. Despite observing a notable performance improvement in the Encyclopedic VQA task, there was a significant drop in performance on the InfoSeek task. Therefore, to maintain the consistency and overall performance of EchoSight, we have decided to continue using Eva-CLIP as our vision backbone.

\begin{table}[htb]
\small
\setlength{\tabcolsep}{3mm}
\begin{tabular}{lcccc}
\toprule
\multirow{2}{*}{Backbone} & \multicolumn{4}{c}{Recall@K}  \\
 & K=1 & K=5 & K=10 & K=20 \\
\midrule
\textbf{Eva-CLIP}  & & & &  \\
\hspace{0.5em}w/o. Reranking  &13.3 &31.3 &41.0 &48.8 \\
\hspace{0.5em}w. Reranking  &\textbf{36.5} &\textbf{47.9} &\textbf{48.8} &48.8 \\
\textbf{DINOv2}  & & & &  \\
\hspace{0.5em}w/o. Reranking  &17.3	&38.6	&46.0	&51.4 \\
\hspace{0.5em}w. Reranking  &\textbf{40.8}	&\textbf{50.7}	&\textbf{51.3}	&51.4 \\
\bottomrule
\end{tabular}
\caption{DINOv2 comparison to CLIP in E-VQA.}
\label{tab:dino_evqa}
\end{table}

\begin{table}[htb]
\small
\setlength{\tabcolsep}{3mm}
\begin{tabular}{lcccc}
\toprule
\multirow{2}{*}{Backbone} & \multicolumn{4}{c}{Recall@K}  \\
 & K=1 & K=5 & K=10 & K=20 \\
\midrule
\textbf{Eva-CLIP}  & & & &  \\
\hspace{0.5em}w/o. Reranking  &45.6 & 67.1 &73.0 &77.9 \\
\hspace{0.5em}w. Reranking  &\textbf{53.2} &\textbf{74.0} &\textbf{77.4} &77.9 \\
\textbf{DINOv2}  & & & &  \\
\hspace{0.5em}w/o. Reranking  &34.7& 53.4	& 60.0	& 64.5 \\
\hspace{0.5em}w. Reranking  &\textbf{38.2}	&\textbf{60.0}	&\textbf{64.1}	&64.5 \\
\bottomrule
\end{tabular}
\caption{DINOv2 comparison to CLIP in InfoSeek.}
\label{tab:dino_info}
\end{table}

\section{Prompt Template}
\subsection{E-VQA}
The prompt we use for LLM when testing E-VQA~\cite{mensink2023encyclopedic} is shown as follow:
\lstset{basicstyle=\ttfamily\footnotesize}
\begin{lstlisting}
USER: Context: <CONTEXT> 
Question: <QUESTION>
The answer is:
\end{lstlisting}

\subsection{InfoSeek}
Due to the strict metrics of exact match are used by InfoSeek~\cite{chen2023can}, we have to consider the format of the prompt so that the generated answer is comparable with the ground truth. Thereby, by using a one-shot example to keep the format correct, our prompt we use for InfoSeek is:
\lstset{basicstyle=\ttfamily\footnotesize, breaklines=true, breakindent=0pt}
\begin{lstlisting}
SYSTEM: You always answer the question the user asks. Do not answer anything else.

USER:Context: The sounthern side of the Alps is next to Lake Como.
Question: Which body of water is this mountain located in or next to?
Just answer the questions, no explanations needed. 
Short answer is: Lake Como

Context: <CONTEXT> 
Question: <QUESTION>
Just answer the questions, no explanations needed. 
Short answer is:
\end{lstlisting}

\end{document}